\documentclass{amia}
\usepackage{graphicx}
\usepackage{times}
\usepackage{hyperref}
\usepackage[labelfont=bf]{caption}
\usepackage[superscript,nomove]{cite}
\usepackage{color}
\usepackage{booktabs}
\usepackage{amssymb}
\usepackage{amsmath}
\usepackage{subfigure} 
\usepackage{bm}
\usepackage{wrapfig}
\begin{document}
\sloppy

\newcommand{\fl}[1]{\textcolor{blue}{\emph{[Fenglong: #1]}}}

\title{Predicting Ulnar Collateral Ligament Injury in Rookie Major League Baseball Pitchers}

\author{Sean A. Rendar$^{1}$, Fenglong Ma, PhD$^{1}$}

\institutes{
    $^1$College of Information Sciences and Technology, Pennsylvania State University, PA, USA\\
}

\maketitle

\section{Introduction}
In the growing world of machine learning and data analytics, scholars are finding new and innovative ways to solve real-world problems. One solution comes by way of an intersection between healthcare, sports statistics, and data science. Within the realm of Major League Baseball (MLB), pitchers are regarded as the most important roster position. They often are among the highest paid players and are crucial to a franchise’s success, but they are more at risk to suffer an injury that sidelines them for over a complete season. The ulnar collateral ligament (UCL) is a small ligament in the elbow that controls the strength and stability of a pitcher’s throwing arm. Due to repetitive strain, it is not uncommon for pitchers to tear it partially or completely during their career. Repairing this injury requires UCL reconstruction surgery, as known informally as Tommy John surgery. In this podium abstract, we want to investigate whether we can use machine learning techniques to predict the UCL injury by analyzing online pitcher data.

There are a multitude of previous related work that focuses on using machine learning methods to predict injuries or the risk of injuries within many different sports~\cite{casals2017sports,huang2021data,jauhiainen2021new,whiteside2016predictors}. For example, Whiteside et al.~\cite{whiteside2016predictors} relied on only \emph{\textbf{104} pitchers} as input information for which they based their significant findings, and used linear regression, naïve bayes and support vector machine (SVM) as the classifier. 
Huang and Jiang~\cite{huang2021data} proposed and defined an artificial intelligence (AI) framework that can be used to develop AI injury prediction solutions across many sports. However, using only \emph{\textbf{21} soccer players} which takes away from the validity of their accuracy measures since there could be bias within such a minimal number of subjects.
Furthermore, Jauhiainen et al. (2020) used the physical data from \emph{\textbf{314} young basketball and floorball players} to train linear regression and random forest to predict injury risk. 

Although existing work can predict the risk of injuries, the small size of dataset may lead to biased predictions. To solve this issue, in this abstract, we aim to create a new, large, and publicly available dataset that is recorded from pitchers’ play to predict the need of Tommy John Surgery. Besides, several machine learning approaches such as K-Nearest Neighbors, Naïve Bayes, and Decision Trees, are used to validate the usability of this new dataset. 

\section{Methods}

\subsection{Dataset Creation}
With the rise of Sabermetrics, the statistical analysis of baseball, it has become quite easy to access player statistics from nearly the beginning of the MLB. The data used for this research comes from Stathead (\url{https://stathead.com/}), a publicly sourced sports almanac that is specialized toward information research across all American professional sports. 

From Stathead, 47 features are recorded on 8,503 pitchers in each of the rookie seasons. Since the first Tommy John procedure took place in 1974, all the pitcher data collected are from after that year, and only contain statistics regarding a pitcher’s rookie reason. These features range from physical features like height, weight, and handedness as well as pitching statistics such as total games played, total inning pitched, hits allowed, runs allowed, and others of that nature. 
The target variable comes from another publicly sourced repository that logs all professional baseball players who have undergone Tommy John (\url{https://bit.ly/3hIY9Ox}). By joining the two datasets by player name, a complete data pool was formed having both the prior discussed features and a binary target of undergoing the surgery or not per pitcher. 

We split the collect datasets into training and testing sets with a ratio 8:2 according to the the year information. Though it is not uncommon for MLB pitchers to face this injury, there is class imbalance among the Tommy John binary classification target. A ratio of nearly 10:1 is observed with there being 7,677 negative cases (no injury) and 826 positive cases (injury occurred).


\subsection{Data Preprocessing and Model Training}
The data are clean and complete for the most part. Any null data points are handled according to what best fit the feature type. That is being replaced with a mean value of a column or a zero value where appropriate. There are also some variables such as the pitchers’ team, if they played in the National or American League, and handedness which are converted into numerical classifiers. Five supervised learning models are explored in this work, including K-Nearest Neighbors (KNN), Naïve Bayes (NB), tree algorithms such as XGboost, Random Forest (RF), and regular Decision Tree (DT), and multiple layer perceptron (MLP).

Since this is an imbalanced classification problem, it requires manipulation in sampling to the training set to increase model learning for the positive class. In this work, we use oversampling of the positive target classification. In addition, grid search cross-validation is used to hyperparameter-tuning. In conjunction, feature selection is also explored due to the high dimensionality of the data most likely being detrimental in both predictive ability and computational efficiency. To do this, K-Best feature selection is used to reduce the features used in modeling from 47 down to 13. This not only raises testing the performance in training and testing, but it also significantly reduces run times that exponentially increases due to cross-validation and the nature of tree-based models that are explored.

\section{Results}
Since the dataset is imbalanced, and we use the Receiver Operating Characteristic/Area Under the Curve (ROCAUC) score as the evaluation metric. Table~\ref{table:ta1} shows the results on different approaches. We can observe that although MLP is a simple deep learning-based approach, it achieves the best performance compared with traditional classification methods. These results show that using advanced deep learning models may be helpful for boosting the performance. 

\begin{table}[!h]
\centering
\begin{tabular}{|c|c|c|c|c|c|c|}
\hline
\textbf{Model} & KNN  & NB & XGboost & RF & DT & MLP\\\hline
\textbf{ROCAUC} & 0.5702 & 0.5463 & 0.6068 &  0.6143& 0.6329 & \textbf{0.6740}  \\
\hline
\end{tabular}
\caption{Result comparison.}
\label{table:ta1}
\end{table}

\section{Discussion}
As we mentioned before, this dataset is significantly imbalanced. To make the two classes balanced, we tried three sampling techniques: oversampling of the positive target classification, undersampling of the negative target classification, and SMOTE sampling which combines the tactics of oversampling and undersampling. However, we found that the oversampling of the positive cases served the best in training models. In the future, we will explore more advanced sampling techniques to further improve the performance.

Though the predictive capability of the models discussed is not as strong as the researchers hoped to have found, a relationship is present. Injury prediction is a complex task and requires a great deal of data manipulation through sampling and feature selection to correctly undertake any form of adequate modeling. It is hoped that this research will serve as a foundation to future progress in this area as a proof of concept has been established. It has powerful implications for not only the MLB, but across all professional sports and maybe even higher-level college athletics.

\makeatletter
\renewcommand{\@biblabel}[1]{\hfill #1.}
\makeatother

\bibliographystyle{vancouver}
\bibliography{sampleBibFile}

\end{document}